\documentclass[letterpaper, 10 pt, conference]{ieeeconf}  
\usepackage{amsfonts}
\usepackage{amssymb}
\usepackage{graphicx}
\usepackage{hyperref}

\usepackage{graphicx}

\IEEEoverridecommandlockouts                              

\usepackage{times} 
\usepackage{amsmath} 

\title{\LARGE \bf
A Reinforcement Learning Based Controller to Minimize Forces on the Crutches of a Lower-Limb Exoskeleton*
}

\author{Aydin Emre Utku$^{1}$, Suzan Ece Ada$^{2}$, Muhammet Hatipoglu$^{2}$, Mustafa Derman, Emre Ugur$^{2}$ and Evren Samur$^{1}$ 
\thanks{*This study is supported by TÜBİTAK (the Scientific and Technological Research Council of Türkiye) under the project no 118E923 and the BİDEB 2210-A program.
}
\thanks{$^{1}$Aydin Emre Utku and Evren Samur are with the Haptics \& Robotics Lab, Department of Mechanical Engineering, Bogazici University, Istanbul, Turkey.  {\tt\small utkuemre.eu@gmail.com, evren.samur@boun.edu.tr.}}%
\thanks{$^{2}$Suzan Ece Ada, Muhammet Hatipoglu, and Emre Ugur are with CoLoRs Lab, Department of Computer Engineering, Bogazici University, Istanbul,
        {\tt\small ece.ada, muhammet.hatipoglu,emre.ugur@boun.edu.tr }}%
}

\begin{document}

\maketitle
\thispagestyle{empty}
\pagestyle{empty}


\begin{abstract}
Metabolic energy consumption of a powered lower-limb exoskeleton user mainly comes from the upper body effort since the lower body is considered to be passive. However, the upper body effort of the users is largely ignored in the literature when designing motion controllers. In this work, we use deep reinforcement learning to develop a locomotion controller that minimizes ground reaction forces (GRF) on crutches. The rationale for minimizing GRF is to reduce the upper body effort of the user. Accordingly, we design a model and a learning framework for a human-exoskeleton system with crutches. We formulate a reward function to encourage the forward displacement of a human-exoskeleton system while satisfying the predetermined constraints of a physical robot. We evaluate our new framework using Proximal Policy Optimization, a state-of-the-art deep reinforcement learning (RL) method, on the MuJoCo physics simulator with different hyperparameters and network architectures over multiple trials. We empirically show that our learning model can generate joint torques based on the joint angle, velocities, and the GRF on the feet and crutch tips. The resulting exoskeleton model can directly generate joint torques from states in line with the RL framework. Finally, we empirically show that policy trained using our method can generate a gait with a 35\% reduction in GRF with respect to the baseline.
\end{abstract}

\section{INTRODUCTION}

Lower-limb exoskeletons, a subclass of legged robots, have a high level of locomotion complexity as human movements are naturally involved. Limited degrees of freedom may oblige the exoskeleton user to use crutches for stable locomotion. Consequently, the upper body activity to control the crutches results in increased metabolic energy consumption. This increased metabolic energy consumption needs to be minimized to improve the comfort of the powered exoskeleton user. The motivation of this study is to fill this gap by developing a controller that takes the crutch movements and contact forces into account to minimize the metabolic energy consumption of an exoskeleton user. 

Even though there are a number of numerical or analytic motion control methods that can solve the problem of generating accurate low-level control commands, they generally require complex mathematical models that describe the dynamic behavior of the whole system. For example, classic PID control is often used in locomotion tasks [1], but we still need to mathematically model the system, especially the contact dynamics to minimize the contact forces. Considering the highly nonlinear nature of this kind of problem, it is a good option to explore the capabilities of deep reinforcement learning (RL) techniques which do not require the adoption of complex expressions of system dynamics. 

Another advantage of deep RL algorithms is improved generalization [2]. Since deep RL algorithms can work with raw data (such as image pixels), the environment does not need to be necessarily defined beforehand. Scaled raw inputs can be supplied to a neural network without any structuring layer for the observation space. Thus, the algorithm can work together with some new set of inputs that were not a part of the training phase.  In deep RL, an agent is defined as a neural network or a combination of neural networks. Some of the architectural design considerations include the number of layers and the number of neurons in each layer, their activation functions, and connectivity structure. These neural networks are designed to map the specific observations to required actions and values. The weight and biases are initialized randomly and then adjusted at every iteration depending on the error, using backpropagation.
 
Deep RL methods are widely used in the literature in the context of prosthetic legs, exoskeletons, and bipedal robots. [3,4,5,6]. Lillicrap et al. [5] demonstrated the generalized efficacy of deep reinforcement learning on different continuous control problems, including multi-joint dynamics and unstable contact dynamics such as cart pole, pendulum swing-up, puck shooting, and robotic manipulation, legged locomotion, and autonomous driving. In their work, Deep Deterministic Policy Gradient (DDPG) is introduced to solve continuous control problems. DDPG uses the Deep Q-Network [6] in the actor-critic framework where the actor generates action, and the critic evaluates.
 
Deep Learning has been used in bipedal robots or lower-body exoskeletons. However, training a policy network such that it directly outputs the desired joint torques using deep RLis also a valid option. Rose et al. [5] used deep neural networks to develop a controller that generates torque values directly. A neural network that generates joint torques based on joint angle, velocity, acceleration, actuator torque, speed, and joint angle goals observations was established and trained by the DDPG algorithm. A generalized reward scheme was used to generate a general reward function that can encompass the successful tracking of goal joint angles and a penalty for exceeding the joint limits. This common reward scheme was summed for all the joints to compute the overall reward. OpenSim-RL, which is an open-source RL environment for multibody physics simulations, was used as the simulation environment. The desired gait pattern was obtained by applying inverse kinematics after using the joint torque profiles available in OpenSim. The learned gait was evaluated after the training, and the desired joint trajectories were observed to be tracked even in the presence of small perturbations. Being able to track the unseen trajectories in training is given as future work.
 
Prior work on learning-based methods for exoskeletons focused on obtaining a desirable gait. However, the upper body effort of the users is largely ignored in these studies. In this work, a state-of-the-art RL algorithm, Proximal Policy Optimization (PPO), with a reward function based on a term that regulates the GRF on the crutch tips is implemented for a lower-limb exoskeleton. The developed RL agents are intended to exhibit a walking behavior with desired gait characteristics and crutch usage. 

The gait characteristics such as linear displacement, lateral displacement, the tilt angle of the human body, and the angle of the feet soles with respect to the ground are kept within an acceptable range while reducing GRF on the crutches of the exoskeleton. By attaining these, the long-term goal of improving the comfort of lower-limb exoskeleton users can be realized.

\begin{table*}[]
\caption{MEASUREMENTS OF THE SELECTED HUMAN SUBJECT [9]} 
\label{t1}
\centering
\begin{tabular}{|p{8mm}|p{8mm}|p{8mm}|p{8mm}|p{8mm}|p{8mm}|p{8mm}|p{8mm}|p{8mm}|p{8mm}|}
\hline Mass [kg] &Height [cm] &Foot size [cm] &Arm span [cm] &Ankle height [cm] &Hip height [cm] &Hip width [cm] &Knee height [cm] &Shoulder width [cm] &Shoulder height [cm]\\
\hline 62.2& 168& 24& 163& 8& 91& 25& 48.5 &35.4 &140\\
\hline
\end{tabular}
\end{table*}

\section{Methodology}
 In this section, the proposed deep RL-based motion controller for an exoskeleton that minimizes the reaction forces exerted on the crutch tips of the users is introduced. We assume that the lower body is completely passive. We first designed a real-to-sim framework based on a real wearable exoskeleton system named Co-Ex [8].

\subsection{Simulation Environment Model}
Deep RL algorithms require large amounts of interaction data for training. Real-world data collection is expensive and dangerous, especially for a wearable exoskeleton that involves humans for testing. Hence, we utilize physics-based simulators to train RL agents. Designing a realistic wearable exoskeleton in the simulation is an integral part of learning locomotion. We use MuJoCo (Multi-Joint Dynamics with Contact) physics simulator [13] to utilize reinforcement learning. Fujimoto et al. [9] showed the efficacy of MuJoCo [13] for different types of locomotion tasks by producing efficient gaits for all the available robot configurations in MuJoCo [13]. In addition to MuJoCo [13], we employ the widely used OpenAI Gym [10] framework for reinforcement learning.

\subsection{Human-Exoskeleton Model}
Although there are not many implementations of a human model with joint actuations, a URDF (Unified Robotics Description Format) model is available [11], in Gazebo (another dynamics simulator used for robotics). The human models in [11] were generated by Human Model Generator [12] using motion capture data collected from 8 different subjects. The links in the human body are modeled as cylinders and boxes in this modeling approach. We have selected subject 2 from the models available in [11] for our study because it has the closest measurements of hip height and knee height to the real exoskeleton designed in [8]. The anthropometric measurements of the selected subject are available in Table I. This model is intended to represent the majority of the degrees of freedom in a human body. To decrease the dimensionality of the model, all joints except ankle, knee, hip, arm, and shoulder joints in the sagittal plane are removed by setting them as fixed joints.
 
\begin{figure}[hbt!]
\centering
\includegraphics[scale=0.2]{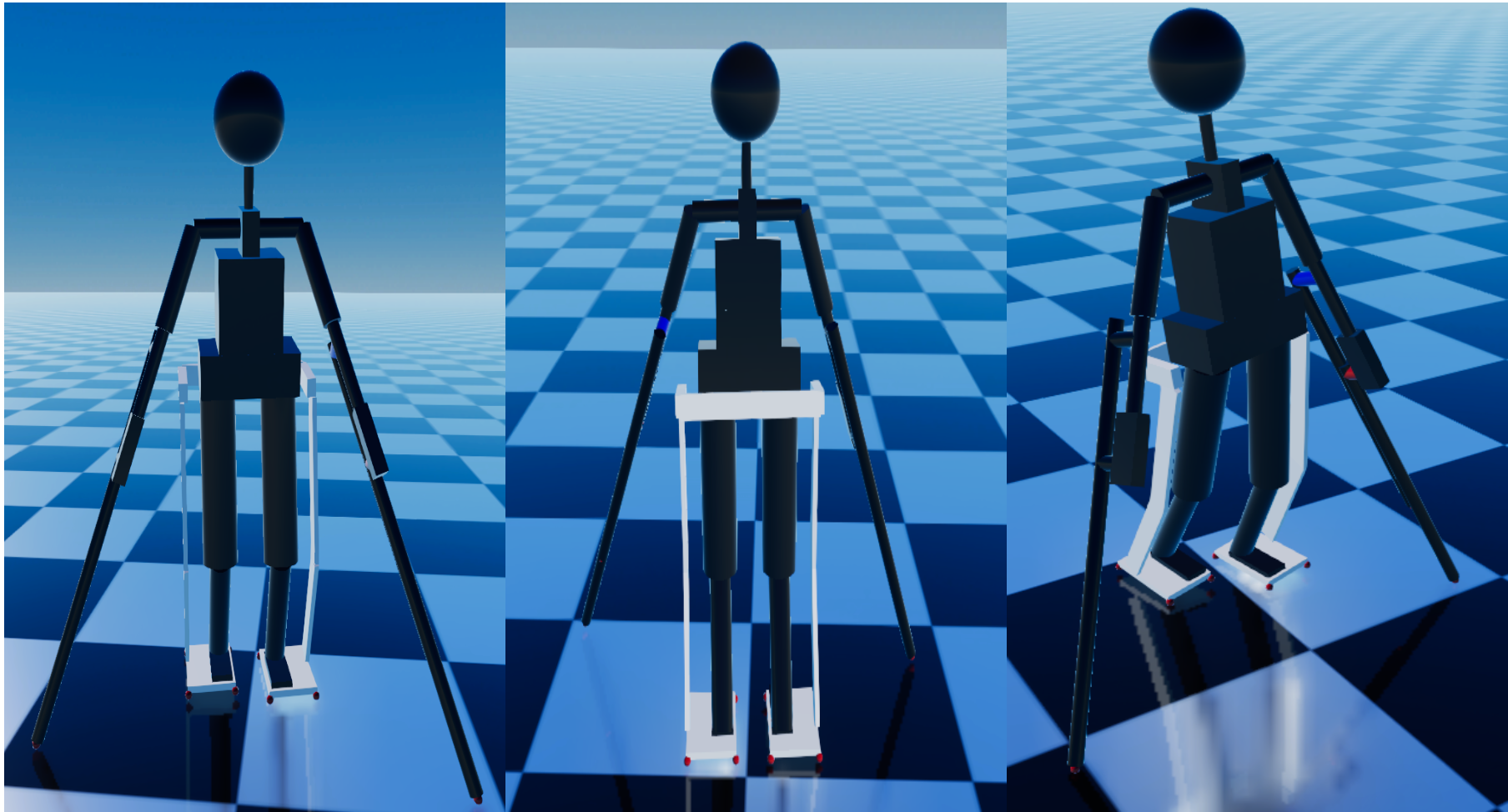}
\caption{Human-exoskeleton system with crutches from different views. (a) Front view (b), Rear view (c), Transverse view.}
\label{fig1}

\end{figure}

\begin{figure}[hbt!]
\centering
\includegraphics[scale=0.2]{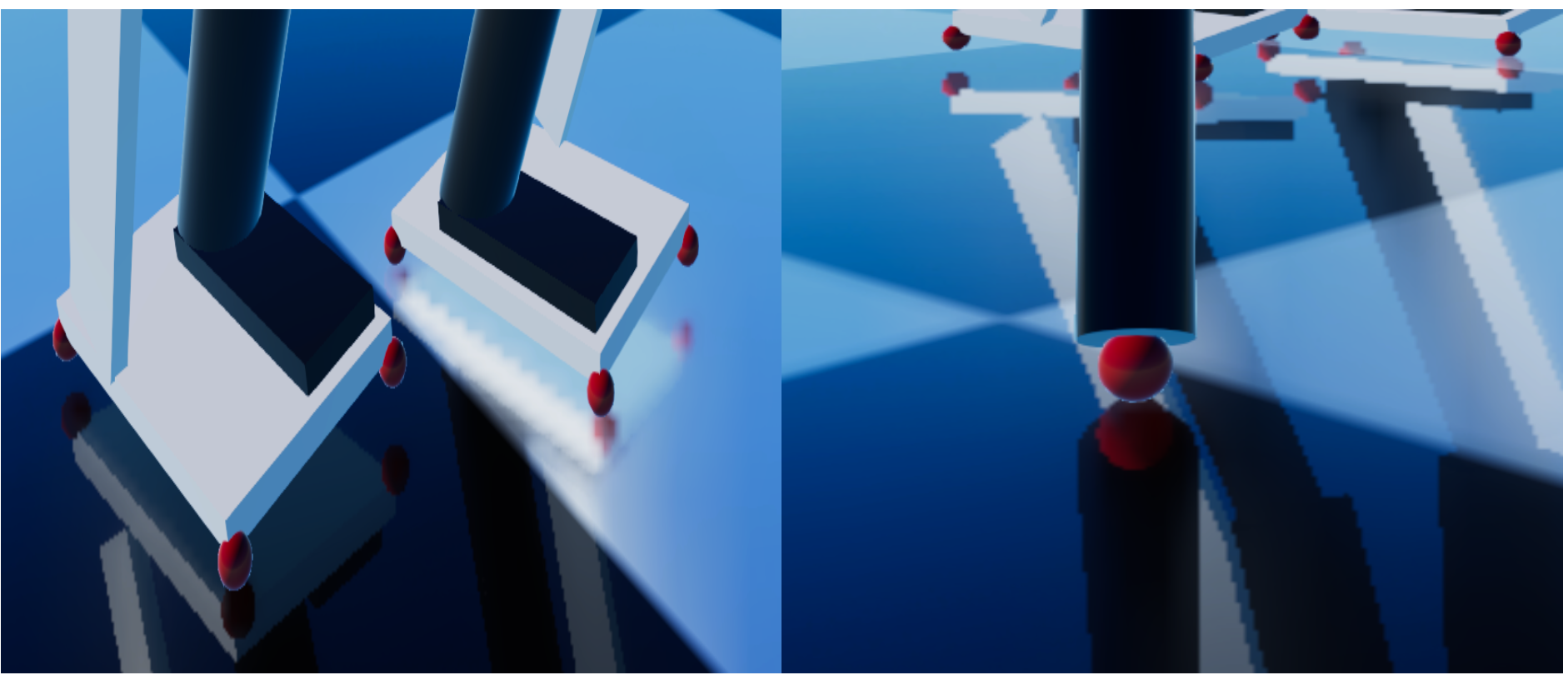}
\caption{Contact spheres. (a) Feet contact, (b) Crutch contact.}
\label{fig2}
\end{figure}

A lower body exoskeleton with crutches is modeled and attached to the human model as illustrated in Figure \ref{fig1}. The main challenge is the correct measurement of the GRF on soles. To collect ground reaction force, data spheres are added to each foot, and crutch tip since MuJoCo does not allow contact force calculation between two planes out of the box. These spheres are modeled as prismatic joints with a predefined stiffness, and thus behave as springs. This contact information is used in constructing the observation vector in reinforcement learning. The contact spheres beneath the feet and the crutches are shown in Figure \ref{fig2}.
 
\subsection{Action and Observation Space in MuJoCo}

Observation space consists of the orientation of the robot base expressed as quaternion, angular positions of ankle, knee, hip, shoulder and arm joints in sagittal plane, and position of contact joints, rate of change of all these variables, actuator forces as well as and velocity. The center of mass (CoM) inertia is also included in this state representation, defined as the body inertia based on CoM in MuJoCo. 

The observation consists of rotation angles with respect to x, y, z axes in quaternion representation, rotation angles of ankle, knee, hip, shoulder and arm joints in sagittal plane, displacements in the prismatic joints that are used to attach the contact spheres, velocities of all these variables, center of mass inertia and velocity and actuator forces. The center of mass inertia is defined as the body inertia based on CoM (center of mass) in MuJoCo. 

   
\subsection{Proximal Policy Optimization}

To realize the mapping between the robot states and required joint torques, PPO [15] is used. PPO uses an actor-critic approach for discrete or continuous action spaces. PPO builds on the Vanilla Policy Gradient with the following advantage function

\begin{equation}
\nabla {J}(\theta)=\hat{{E}}_{{t}}\left[\nabla_\theta \log \pi_\theta\left({a}_{{t}} \mid {s}_{{t}}\right) \hat{{A}}_{{t}}\right]
\end{equation}  

where $\pi_\theta$ is the stochastic policy that outputs the mean and variance of the probability distribution over actions, $\hat{{A}}_t$ is the estimation of advantage at time $t$ and $\hat{{E}}_{{t}}$ is the expectation at time t. To alleviate the large gradient updates, a clipped surrogate objective is introduced:
\begin{equation}
{L}^{{CLIP}}(\theta)=\hat{{E}}_t\left[\min \left(r_t(\theta) \hat{{A}}_t, {clip}\left(r_t(\theta), 1-\varepsilon, 1+\varepsilon\right) \hat{{A}}_t\right)\right]
\end{equation}  
where ${r_t}$ is the ratio

\begin{equation}
\mathbf{r}_{(\mathrm{t})}(\theta)=\frac{\pi_\theta\left(\mathrm{a}_{\mathrm{t}} \mid \mathrm{s}_{\mathrm{t}}\right)}{\pi_{\theta, \mathrm{old}}\left(\mathrm{a}_{\mathrm{t}} \mid \mathrm{s}_{\mathrm{t}}\right)}
\end{equation}  

and "clip" is the operation of clipping the first argument $r_t(\theta)$ within a range $[1-\varepsilon, 1+\varepsilon]$.

\begin{figure}[thpb]
\centering
\includegraphics[scale=0.2]{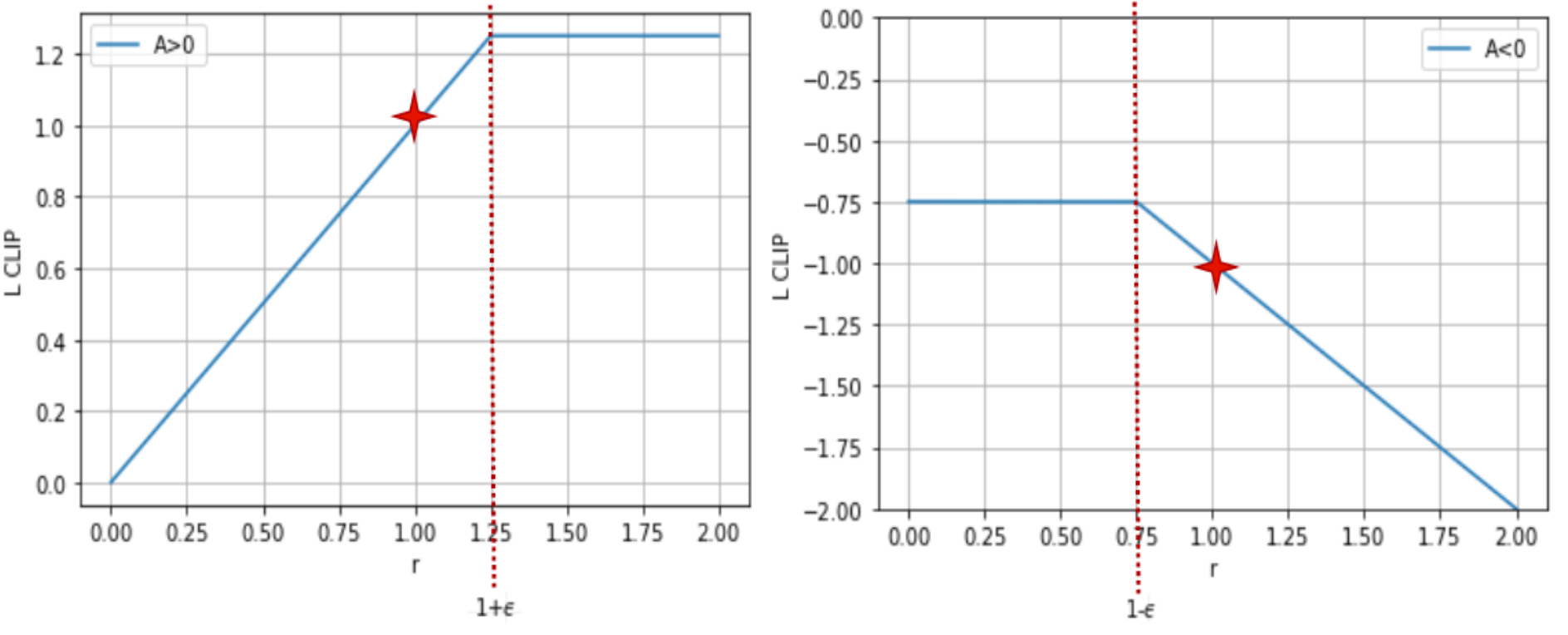}
\caption{Behavior of $L^{CLIP}$ with $\varepsilon=0.25$ for $(a)A>0,(b)A<0$.}
\label{fig3}
\end{figure}

The behavior of $L^{C L I P}$ is shown in Figure \ref{fig3}. When the advantage is negative, ratios less than " $1-\epsilon$ " are clipped to decrease the likelihood of the action under the new policy. Likewise, when the advantage is positive, ratios above " $1+\epsilon$ " are clipped to constrain large updates.
The final entropy regularized actor-critic loss is
\begin{equation}
\hat{E}_t\left[{~L}^{CLIP}_t{(\theta)-c_1 {~L}_t}^{{VF}}(\theta)+{c}_2 S\left[\pi_\theta\right]\left(s_t\right)\right]
\end{equation}  
where $c_1$ and $c_2$ are weighting coefficients, $S$ is entropy bonus. The critic loss $L_t^{V F}(\theta)$ is 

\begin{equation}
{L}_{{t}}^{{VF}}(\theta)=\left({V}_\theta\left({s}_{{t}}\right)-{V}_{{t}}^{{targ}}\right)^2
\end{equation}  

where the ${V}_\theta\left({s}_{{t}}\right)$ is the critic network and  ${V}_{{t}}^{{targ}}$ is the target value function. The stochastic policy $\pi_\theta$ is generally modeled as probabilistic Gaussian or beta distributions (Chou et al. [16]). In this study, multivariate Gaussian distribution with diagonal covariance is used as the output of the policy network following the implementation in OpenAI Baselines [17]

\subsection{Reward Shaping}

The reward function is designed to promote the human-exoskeleton system walk in a straight posture, without deviating to left or right directions, tilting the body beyond constraints and falling. The terms that shape the reward function are $r_{\text{walk}}$, $r_{\text{walk\_straight}}$, $r_{\text{dont\_fall}}$ $r_{\text{action}}$, and $r_{\text{orientation}}$. After obtaining the desired locomotion behavior, additional terms to optimize the gait are added. The terms that shape the reward function are $r_{\text{walk}}$, $r_{\text{walk\_straight}}$, $r_{\text{dont\_fall}}$, $r_{\text{action}}$, $r_{\text{orientation}}$, $r_{\text{flat\_contact}}$, $r_{\text{crutch\_reaction\_force}}$, $r_{\text{hip\_angle}}$ and $r_{\text{ensure\_crutch\_contact}}$.

The walking reward encourages the human-exoskeleton system to translate its root position in the forward direction. It is defined as
\begin{equation}
r_{\text{walk}}=e^{-{c}_{{walk}}\left(\dot{{p}}_{{x}}-\dot{{p}}_{\text{x\_des}}\right)^2}
\end{equation}  

where $\dot{{p}}_{{x}}$ is the derivative of the CoM position in ${x}$ coordinate of the current state, $\dot{{p}}_{\text{x\_des}}$ is the desired forward velocity and it is set as $0.25 {~m} / {s}$, and $c_{walk}$ is a hyperparameter set to $5 \times 10^5$.

The walk linear reward discourages the system from walking to the left or right side

\begin{equation}
r_{\text{walk\_straight }}=-\left|p_y\right|
\end{equation}

where $p_y$ is the position in the y-coordinate.
Do not fall reward keeps the height of the root within an acceptable height range.

\begin{equation}
r_{\text{dontfall }}=\left\{\begin{array}{cl}
5, & \text{if} p_z^{min}<p_z<p_z^{max}\\
0, & \text{ else }
\end{array}\right.
\end{equation}
$p_z$ is the position in z coordinate of current state, $p_z^{min}$ and $p_z^{max}$ are chosen as 0.65 and 3.0 respectively.

The action reward discourages the exoskeleton from generating exaggerated actions where ${F}$ is the 6-D vector composed of joint torques proposed by the policy network.

\begin{equation}
{r}_{\text{orientation }}=-8\left({a}_{{z}}-0.35\right)^2
\end{equation}

The orientation reward is used to discourage tilting beyond constraints, where $a_z$ is the angle of the body to the forward direction. $0.35 {rad}$ is selected based on the observations done on exoskeleton users.

\begin{equation}
{r}_{\text{flat\_contact }}=-10\left({r}_{\text{foot\_flat }}+{l}_{\text{foot\_flat }}\right)^2
\end{equation}
The flat contact reward is used to keep the feet parallel to the ground to avoid tripping, where
\begin{equation}
{r}_{\text{foot\_flat}}=\left({a}_{{z}}+{a}_{\text{right\_hip }}+{a}_{\text{right\_knee }}+{a}_{\text{right\_ankle }}\right)^2  
\end{equation}

\begin{equation}
l_{\text{foot\_flat}}=\left({a}_{{z}}+{a}_{\text{left\_hip}}+{a}_{\text{left\_knee }}+{a}_{\text{left\_ankle}}\right)^2
\end{equation}

and ${a}_{{x}_{-} {y}}$ is the angle of related foot-joint combination.

The crutch reaction force $r_{\text{crutch\_reaction\_force }}$ reward discourages the human exoskeleton system from having too much ground reaction force on the crutches:
\begin{equation}
-{w}_{\text{crutch\_reaction\_force }}\left({d}_{\text{crutch\_r}}+{d}_{\text{crutch\_l}}\right)
\end{equation}

where ${d}_{\text{crutch\_r}}$ and ${d}_{\text{crutch\_l}}$ are the displacements on the tip of the right crutch and left crutch, respectively. These displacements represent the ground reaction forces on the crutch tips.

\begin{equation}
r_{\text{hip\_angle}}=\left\{\begin{array}{cl}
-2,\text{if } a^{\text{hip\_r}}<0 \text{ and } a^{\text{hip\_l}}<0 \\
0,\text{else}
\end{array}\right.
\end{equation}

Hip angle reward discourages the exoskeleton to extend the hip angle backwards with respect to the upper body as the hip angle always stays positive during a natural looking lower body exoskeleton gait, where $a^{\text{hip\_r}}$ and $a^{\text{hip\_l}}<0$ are the angle of right hip and left hip, respectively.

Ensure crutch contact reward prevents the human exoskeleton system to stop the contact of the two crutches on the ground at the same time.

\begin{equation}
r_{\text{ensure\_crutch\_contact}}=\left\{\begin{array}{cl}
-2, & \text{ if } {d}_{\text{crutch\_r}}<0.003 \text{ and }\\& {d}_{\text{crutch\_l}}<0.003 \\
0, & \text{ else }
\end{array}\right.
\end{equation}

where ${d}_{\text{crutch\_r}}$ and ${d}_{\text{crutch\_l}}$ are the displacements on the tip of the right crutch and left crutch, respectively. The value 0.003 is selected as the threshold to infer the contact information with the ground.

The final reward function can be defined as
\begin{equation}
{r}=\begin{array}{cl}{r}_{\text{walk}}+{r}_{\text{walk\_straight}}+{r}_{\text{dont\_fall }}\\
+{r}_{\text{action}}+{r}_{\text{orientation}} \\
 +{r}_{\text{flat\_contact }}+{r}_{\text{crutch\_reaction\_force}}+{r}_{\text{hip\_angle }} \\
 +{r}_{\text{ensure\_crutch\_contact }} .
 \end{array}
 \end{equation}
 
\subsection{Overall Implementation}
The URDF file that models the system is transformed into XML format to use in MuJoCo for RL. (18) is used as the reward function. PPO and actor-critic network hyperparameters are provided in Table \ref{t2} and Table \ref{t3}, respectively.

Different  weights used for GRF on the crutch tips are provided in Table IV. For each, training results using five different seeds and four different RL agents are obtained over 8000 iterations. The proposed reward function with GRF minimization term is evaluated with a baseline PPO, which does not have the ground reaction force loss signal in its reward function. Each experiment is done with the same set of random seeds five times and, more importantly, with the same network and PPO hyperparameters.

\begin{table*}[h]
\caption{PPO HYPERPARAMETERS} 
\label{t2}
\centering
\begin{tabular}{|c|c|c|c|c|c|c|c|}
\hline & $\boldsymbol{\varepsilon}$ & ${K}_{\text{epochs }}$ & $\mathbf{L}_{\mathbf{a}}$ & $\mathbf{L}_{\mathbf{c}}$ & $\begin{array}{c}\text{ Entropy } \\
\text{Coefficient }\end{array}$ & $\begin{array}{c}\text{ Entropy } \\
\text{Coefficient } \\
\text{Decay }\end{array}$ & $\boldsymbol{\lambda}$ \\
\hline Value & 0.2 & 10 & 64 & 64 & $1 {e}-3$ & 0.99 & 0.95 \\
\hline
\end{tabular}
\end{table*}

\begin{table}[b]
\caption{NETWORK PROPERTIES} 
\label{t3}
\centering
\begin{tabular}{|c|c|c|}
\hline Network &Actor & Critic\\
\hline Number of Hidden Layers & 1 & 1\\
\hline Hidden  Layer Width & 200&200 \\
\hline Input Layer Activation& Tanh& Tanh \\
\hline Hidden Layer Activation& Tanh &Tanh\\
\hline Output Layer Activation & Softplus&Linear\\
\hline
\end{tabular}
\end{table}

\begin{table}[hbt!]
\caption{CRUTCH REACTION FORCE WEIGHTS} 
\label{t4}
\begin{center}
\begin{tabular}{|c|c|}
\hline RL Agent No & $w_{\text{crutch\_reaction\_force }}$ \\
\hline 1 & 40000 \\
\hline 2 & 30000 \\
\hline 3 & 20000 \\
\hline 4 & 10000 \\
\hline
\end{tabular}
\end{center}
\end{table}

\section{Results}
\begin{figure*}[h]
\centering
\includegraphics[scale=0.35]{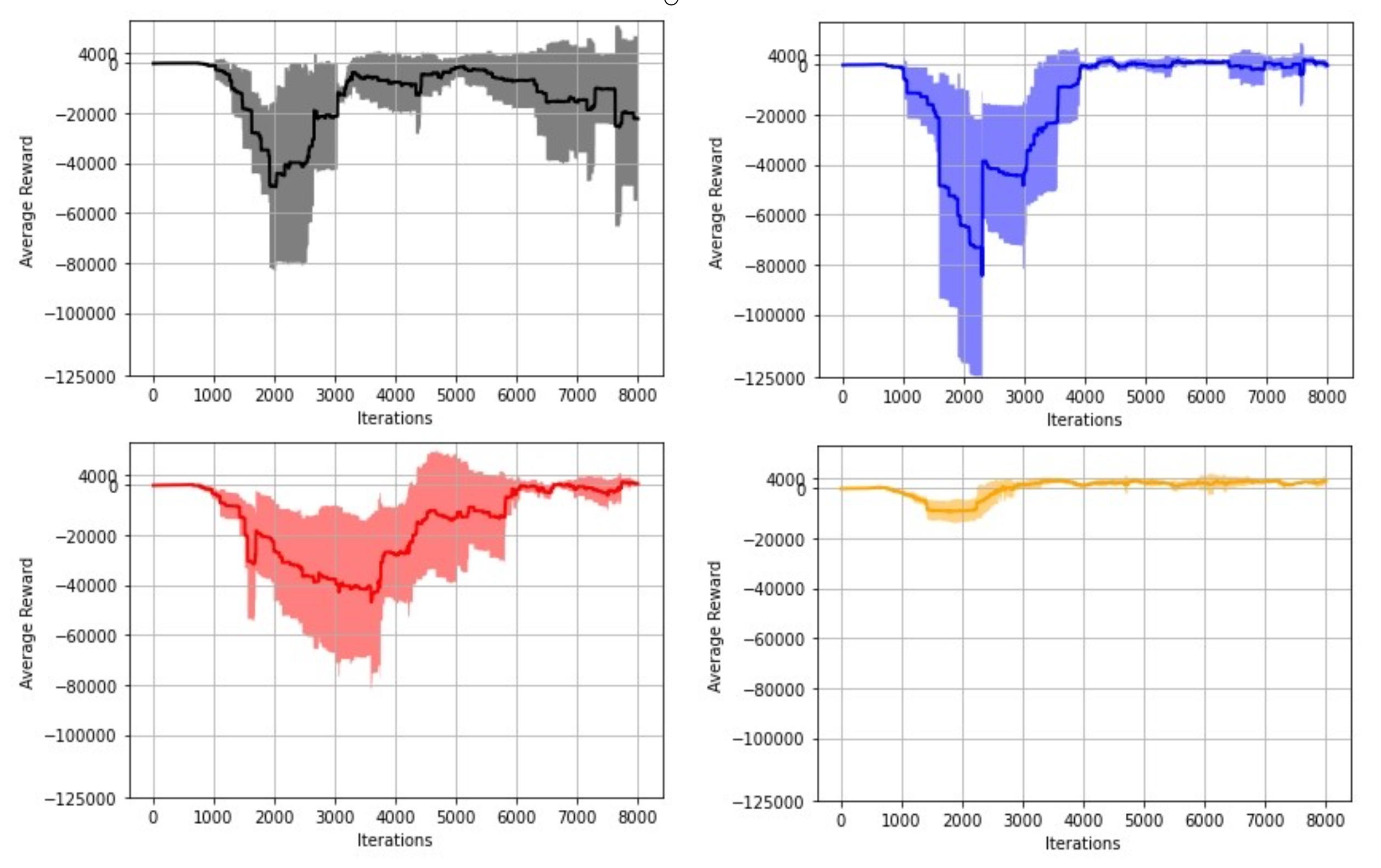}
 \caption{Average cumulative returns of different learning experiments for a range of values for $w_{crutch\_reaction\_force}$. $w_{crutch\_reaction\_force}$ = 4x104 on top-left, 3x104 on top-right, 2x104 on bottom-left, 1x104 on bottom-right.}
 \label{fig4}
\end{figure*}

The algorithm and the reward function were used together to train the human-exoskeleton system with crutches. Training experiments were carried out for different values for $w_{crutch\_reaction\_force}$. To evaluate the overall training regime and to compare different learned walking patterns, the number of training iterations was kept at 8000 throughout the experiments. The expected cumulative rewards during training over four different $w_{crutch\_reaction\_force}$ values are depicted in Figure \ref{fig4}. The curves were smoothed with a moving average filter with a window size of 100. 
For this task, it is empirically found that reaching a cumulative reward above 2000 results in a controller capable of tracking target speed while satisfying hard constraints such as position and orientation limits for the robot CoM. Hence, it is found that increasing the crutch cost beyond 3x104 leads to an unstable learning regime with high variance. In contrast, the agent learns to collect rewards greater than 2000 if the crutch cost coefficient is kept below 3x104.  Experiments without this term were also carried out to assess the contribution of the crutch cost to learning the walking pattern. Since the crutch cost and cumulative return depend on the loss coefficient itself, the learned policies of the experiments were evaluated with the same set of cost coefficients. An overview of this comparative analysis is presented in Table V, where mean crutch cost calculated with the same weight, absolute deviation ratio from the target linear velocity and root orientation, as well as mean lateral displacement, are summarized. 

The mean crutch reaction cost was calculated based on (15) with ${w}_{\text{crutch\_reaction\_force}}=4\times10^4$ for all RL agents at each timestep. The average of these values at each timestep was calculated and given as mean crutch reaction cost in Table V. Note that ${w}_{\text{crutch\_reaction\_force}}$ was selected as $4\times10^4$ only during the testing in order to compare the GRFs of different RL agents, because we are interested in comparing the term $({d}_{\text{crutch\_r}}^2+{d}_{\text{crutch\_l}}^2)$ in (15) for different RL agents.

To calculate the mean absolute percentage error for linear velocity in Table V, CoM velocities were subtracted from the desired velocity, which was defined as $0.25 {~m} / {s}$ for each timestep, and absolute values of these subtractions were summed up over 2000 timesteps. Then these integrated values were divided by 2000 to calculate the mean absolute percentage error for linear velocity. The mean absolute percentage error for root orientation was been calculated by the same method as used to calculate the mean absolute velocity error. In order to calculate the mean absolute lateral displacement, the absolute values of lateral displacements in each timestep were summed up, and the sum was divided by 2000.

\begin{figure}[hbt!]
\centering
\includegraphics[scale=0.25]{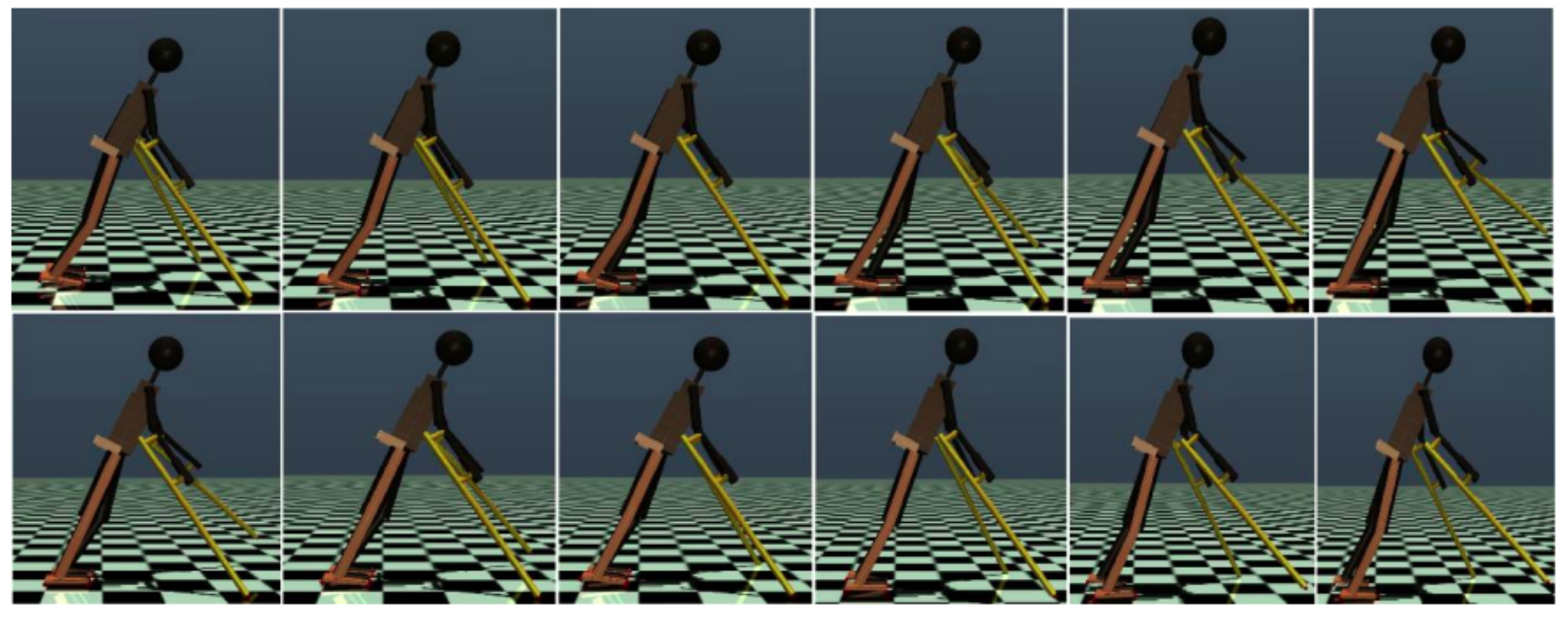}
\caption{A snippet of the gait sequence of the agent in Experiment 1}
\label{figurelabel5}
\end{figure}

The learned walking patterns show notable improvements in terms of crutch reaction forces during walking. Although increasing the weight of this cost term leads to an unstable training regime, it results in the least crutch reaction cost and shows good performance in tracking the reference velocity. A snippet of the gait sequence of this agent can be seen in Figure \ref{figurelabel5}.

\begin{table*}[hbt!]
\caption{COMPARISON OF DIFFERENT AGENTS IN TERMS OF A SUBSET OF REWARD COMPONENTS} 
\label{t5}
\centering
\begin{tabular}{|c|c|c|c|c|}
\hline Experiment & $\begin{array}{c} \text{ Mean crutch } \\
\text{ reaction cost }\end{array}$ & $\begin{array}{c} \text{ Mean absolute } \\
\text{ percentage error } \\
\text{ for linear velocity } \\
\text{ (\%) }\end{array}$ & $\begin{array}{c}\text{ Mean absolute } \\
\text{ percentage error} \\
\text{for root orientation} \\
\text{ (\%) }\end{array}$ & $\begin{array}{c}\text{ Mean } \\
\text{ absolute } \\
\text{ displacement } \\
\text{ (m) }\end{array}$ \\
\hline No crutch loss & 0.769 & 17.10 & 16.71 & 0.03 \\
\hline 1 & 0.501 & 11.11 & 34.96 & 0.07 \\
\hline 2 & 0.588 & 17.26 & 12.19 & 0.17 \\
\hline 3 & 0.619 & 16.78 & 22.51 & 0.07 \\
\hline 4 & 0.665 & 15.58 & 23.82 & 0.05 \\
\hline
\end{tabular}
\end{table*}

\section{CONCLUSIONS}
 This paper introduces a learning-based approach for minimizing forces on crutches, hence the upper-body fatigue, of a user of an exoskeleton that helps with walking. A joint human-exoskeleton model has been constructed in MuJoCo simulation environment utilizing PPO, a state-of-the-art actor-critic reinforcement learning algorithm, with a custom reward function for a real lower-limb exoskeleton. As a result, a range of successful walking behaviors was learned. It is shown by a set of experiments that adding a loss term to the reward signal that is proportional to the ground reaction force exerted on the crutches minimized these forces. 

\section*{ACKNOWLEDGMENT}
We thank Barkan Uğurlu and Özkan Bebek from Özyeğin University who provided invaluable insight and expertise that significantly assisted the research.


\end{document}